\documentclass[twocolumn,10pt]{IEEEtran}
\usepackage{flushend}
\usepackage{amsmath}
\usepackage{flushend}
\usepackage{multirow}
\usepackage{color}
\usepackage{enumerate}
\usepackage{easylist}
\usepackage{tikz}
\usetikzlibrary{shapes,arrows,shadows}
\usepackage{times}
\usepackage[symbol]{footmisc}
\usepackage{verbatim}
\usepackage{amsthm}
\usepackage{amssymb }
\usepackage{amsmath}
\usepackage{bm}
\usepackage{cite}
\usepackage{float}
\usepackage{mathrsfs}
\usepackage{amsfonts}
\usepackage{graphicx}
\usepackage{caption}
\usepackage{algorithm, algpseudocode}
\usepackage{subcaption}
\usepackage{footnote}
\usepackage{footmisc}
\usepackage[colorinlistoftodos]{todonotes}
\usepackage{hyperref}
\usepackage{color,soul}

\def\EM{\text{FS-HGR}}
\def\ND{N_D}
\def\NS{N_S}
\def\i{_{i}}
\def\j{_{j}}
\def\x{\bm{x}}

\algnewcommand\Input{\item[\hspace{6pt}\textbf{Input:}]}
\algnewcommand\Output{\item[\hspace{6pt}\textbf{Output:}]}
\algnewcommand\OutputVal{\textbf{output} }
\def\x{{\mathbf x}}

\def\minmax{\text{Minmax}}
\title{$\EM$: Few-shot Learning for Hand Gesture Recognition via ElectroMyography\thanks{This Project was partially supported by the Department of National Defence’s Innovation for Defence Excellence and Security (IDEaS)
program, Canada.}}

\author{Elahe Rahimian$^{\dagger}$, \textit{Student Member, IEEE}, Soheil Zabihi$^{\mathsection}$, \textit{Student Member, IEEE}, Amir Asif$^{\dagger\dagger}$, \textit{Senior Member, IEEE}, Dario Farina$^{\ddagger\ddagger}$, \textit{Fellow, IEEE}, Seyed Farokh Atashzar$^{\ddagger}$, \textit{Member, IEEE}, and Arash Mohammadi$\dagger$, \textit{Senior Member, IEEE} \\
$^{\dagger}$Concordia Institute for Information System Engineering (CIISE), Concordia University, Montreal, QC, Canada\\
$^{\mathsection}$Electrical and Computer Engineering, Concordia University, Montreal, QC, Canada\\
$^{\ddagger\ddagger}$ Department of Bioengineering, Imperial College London, London, UK \\
$^{\ddagger}$ Electrical \& Computer Engineering and Mechanical \& Aerospace Engineering, New York University, USA\\
$^{\dagger\dagger}$Electrical and Computer Engineering, York University, Toronto, Canada
}

\begin{document}

\date{\today}
\maketitle
\thispagestyle{empty}

\begin{abstract}
This work is motivated by the recent advances in Deep Neural Networks (DNNs) and their widespread applications in human-machine interfaces. DNNs have been recently used for detecting the intended hand gesture through processing of surface electromyogram (sEMG) signals. The ultimate goal of these approaches is to realize high-performance controllers for prosthetic. However, although DNNs have shown superior accuracy than conventional methods when large amounts of data are available for training, their performance substantially decreases when data are limited. Collecting large datasets for training may be feasible in research laboratories, but it is not a practical approach for real-life applications. Therefore, there is an unmet need for the design of a modern gesture detection technique that relies on minimal training data while providing high accuracy. Here we propose an innovative and novel ``Few-Shot Learning'' framework based on the formulation of meta-learning, referred to as the $\EM$, to address this need. Few-shot learning is a variant of domain adaptation with the goal of inferring the required output based on just one or a few training examples. More specifically, the proposed $\EM$ quickly generalizes after seeing very few examples from each class. The proposed approach led to $85.94\%$ classification accuracy on \textit{new repetitions with few-shot observation ($5$-way $5$-shot)}, $81.29\%$ accuracy on \textit{new subjects with few-shot observation ($5$-way $5$-shot)}, and $73.36\%$ accuracy on \textit{new gestures with few-shot observation ($5$-way $5$-shot)}.
\end{abstract}
\section{Introduction}\label{sec:Introduction}
The recent advances in Machine Learning (ML) and Deep Neural Networks (DNNs) coupled with innovations in rehabilitation technologies have resulted in a surge of significant interest in the development of advanced myoelectric prosthesis control systems. Hand motion recognition via sEMG signals~\cite{2_Dario, Dario} is considered as a central approach in the literature. Conventional ML techniques, such as Linear Discriminant Analysis (LDA)~\cite{LDA-SVM, SVM, LDA} and Support Vector Machines (SVMs)~\cite{LDA-SVM, SVM, DB5}, have been used for detecting the intended hand gesture through processing of surface EMG (sEMG) signals. Although classical pattern-recognition-based myoelectric control has been widely studied in academic settings over the last decades, the advanced methodologies have not been used in many commercial examples. This is due to a noticeable gap~\cite{2_Dario, 19-Patrick3, 3_Dario} between real-world challenges and existing methodologies. Among the reasons for this gap are:
\begin{itemize}
\item[(i)] \textit{Training Time}: The first problem is the extended training time required by the end-user to mitigate the differences between the desired and performed movements. Such a training process, which is time consuming, tedious and unpleasant, can take up to several days in practice.
\item[(ii)] \textit{Variability in the characteristics of sEMG Signals}: The second issue is the variability in the nature of the sEMG signals. This variability is caused by (a) Time-dependent and stochastic nature of the neural drive to muscles; (b) Dependency of the neural drive to the dynamic and kinematics of tasks, and; (c) Variability in neural control strategies between different users and the changes caused by amputations. In addition, sEMG recording could vary based on electrode location. Given such variations, therefore, the probability distributions of sEMG signals may be different over time. Consequently, models trained based on some specific observations may not consistently and directly be reused over time. This would require retraining and recalibration,  which cannot be done often in real-life applications.
\end{itemize}
Recently, DNNs have been designed and used by our team~\cite{FD-TBME, JMRR_Elahe, Globalsip_Elahe, Icassp_Elahe} and other research groups~\cite{Josephs:2020, sensor2020, Peng:2020,  pattern_letter2019, WeiNet,YuNet}, for myocontrol, achieving superior classification performance than conventional approaches. However, DNNs need large training data to achieve high performance. This may be feasible in laboratory conditions but poses constraints in the practical use of prostheses in real-life applications. There is an unmet need for the design of a modern gesture detection technique that relies on minimal training data while achieving high performance.

In this paper we introduce, for the first time, the concept of few-shot training for myoelectric systems. Few-shot learning minimizes the need for recalibration and would allow the user to retrain the ML core of control, by only few basic exercises instead of extensive recalibration procedures. For this purpose, here we propose an innovative \textit{Few-Shot Learning framework, referred to as the $\EM$}\footnote{The source code of the proposed $\EM$ framework is available at: https://ellarahimian.github.io/FS-HGR/}. The proposed meta-learning $\EM$ architecture takes advantage of domain knowledge and requires a small amount of training data (when compared with traditional counterparts) to decode new gestures of the same or new users. The paper makes the following contributions:
\begin{itemize}
\item A class of architectures is introduced for sEMG meta-learning, where the meta-learner, via adaptation, quickly incorporates and refers to the experience based on just few training examples.
\item The proposed $\EM$ framework allows a myoelectric controller that has been built based on background data to adapt to the changes in the stochastic characteristics of sEMG signals. The adaptation can be achieved with a small number of new observations making it suitable for clinical implementations and practical applications.
\item By proposing the $\EM$ framework, which utilizes a combination of temporal convolutions and attention mechanisms, we provide a novel venue for adopting few-shot learning, to not only reduce the training time, but also to eventually mitigate the significant challenge of  variability in the characteristics of sEMG signals.
\end{itemize}
The paper is organized as follows: Section~\ref{sec:RelWorks} provides a brief overview of relevant literature. In Section~\ref{sec:database}, we present the dataset used in development of the proposed $\EM$ framework together with the pre-processing step. The proposed $\EM$ architecture is developed in Section~\ref{sec:model}. Experimental results and different evaluation scenarios are presented in Section~\ref{sec:results}. Finally, Section~\ref{sec:con} concludes the~paper.

\section{Related Works}\label{sec:RelWorks}
A common strategy used for hand gesture recognition in recent works is applying DNN with the focus on improving hand gestures classification performance on ``\textit{never-seen-before repetitions}''. Along this line of research, several state-of-the-art works~\cite{Icassp_Elahe, JMRR_Elahe, Globalsip_Elahe, sensor2020, pattern_letter2019,WeiNet, YuNet, DingNet, ZhaiNet, GengNet, AtzoriNet} mainly used the Ninapro database~\cite{1_Ninapro, 2_Ninapro, 3_Ninapro}, which is a public dataset providing kinematic and sEMG signals from $52$ finger, hand, and wrist movements. The Ninapro database is similar to data obtained in real-world conditions, and as such it allows development of advanced DNN-based recognition frameworks.

The common approach in recent studies~\cite{Icassp_Elahe, JMRR_Elahe, Globalsip_Elahe, sensor2020, pattern_letter2019,WeiNet, YuNet, DingNet, ZhaiNet, GengNet, AtzoriNet}, following the recommendations provided by the Ninapro database, is to train DNN-based models on a training set consisting of approximately $2/3$ of the gesture trials of each subject.   The evaluation is then performed on the remaining trials constituting the test set. Although existing DNN techniques achieve promising performance on never-seen-before repetitions, they fail to function properly if the repetition is not extensively explored~\cite{hugo, finn, snail}. Thus, for a new user or a new gesture, a significant amount of training should be conducted and the whole learning process should be redone, assuming  a small variation between the new class and the previous classes. If the aforementioned change is more than minimal, there may be the need to recalibrate the whole process for all classes. In addition, existing DNN-based methodologies require large training datasets and perform poorly on tasks with only a few examples being available for training purposes.

In Reference~\cite{12}, the authors proposed a domain adaptation method that maps both the original and target data into a common domain, while keeping the topology of the input data probability distributions. For this purpose, the authors used a local dataset, where the sEMG data was acquired by repetitive gripping tasks while data was collected from $8$ subjects. In addition to the above, Transfer Learning (TL) was also used to adopt a pre-trained model and leverage the knowledge acquired from multiple subjects and speed up the training process for the new users. In~\cite{transfer-learning, transfer-learning2}, the authors proposed a TL-based algorithm adopting Convolutional Neural Networks (CNN) to transfer knowledge across multiple subjects for sEMG-based hand gesture recognition. The authors in~\cite{transfer-learning, transfer-learning2}, applied the Myo armband to collect sEMG signals and used the fifth Ninapro database, which contains data from $10$ intact-limb subjects.
The pre-training for each participant was done employing the training sets of the remaining nine participants and the average accuracy was obtained over the 10 participants of the Ninapro DB5~\cite{DB5}.
Finally, References~\cite{4_Dario, inter-session} applied deep learning along with domain adaptation techniques for inter-session classification to improve the robustness for the long-term uses. Due to the variability of the signal space, the generalizability of existing techniques is questionable and it is not clear how they would perform in real-life scenarios when the training data is limited and extensive data collection cannot be done with high frequency to capture the changes.

In summary, there is an urgent need to develop adaptive learning methods with the focus on designing a classifier which can be adopted for new subjects based on only a few examples through a fast learning approach. This is a challenging task since many factors, such as electrode location and muscle fiber lengthening/shortening, can affect the collected sEMG signals. Moreover, the differences between users and the changes caused by amputations result in discrepancies between different conditions~\cite{3_Dario, Dario}. To the best of our knowledge, this is the first time that \textit{Few-shot Learning} is adopted in the literature to classify $49$ hand gestures  on \textit{new} subjects using only a small (one to five) number of training examples.

\section{Material and Methods}\label{sec:database}
\subsection{Database}
The proposed $\EM$ architecture was evaluated on the Ninapro~\cite{1_Ninapro, 2_Ninapro, 3_Ninapro} benchmark database, which is a publicly available dataset for hand gesture recognition tasks. Ninapro is a widely used benchmark for evaluation of different models developed using sparse multichannel sEMG signals.

In this work, the second Ninapro database~\cite{1_Ninapro} referred to as the DB2 was utilized. Delsys Trigno Wireless EMG system with $12$ wireless electrodes (channels) was used in the DB2 dataset to collect electrical activities of muscles at a rate of $2$ kHz. The dataset consists of signals collected from $28$ men and $12$ women with age $29.9 \pm 3.9$ years, among whom $34$ are right-handed and $6$ are left-handed. The DB2 consists of $50$ gestures including wrist, hand, grasping, and functional movements along with force patterns from $40$ healthy (intact-limb) subjects. The subjects repeated each movement $6$ times, each time lasted for $5$ seconds followed by $3$ seconds of rest. More detail on the Ninapro database are described in Reference~\cite{1_Ninapro}.

\subsection{Pre-processing Step}
Following the pre-processing procedure established in previous studies~\cite{GengNet, pattern_letter2019, 1_Ninapro, AtzoriNet}, we used a $1^{\text{st}}$ order low-pass Butterworth filter to smooth the electrical activities of muscles. Moreover, we applied \textit{$\mu$-law} transformation to magnify the output of sensors with small magnitude (in a logarithmic fashion), while keeping the scale of those sensors having larger values over time. This transformation approach has been used traditionally in speech and communication domains for quantization purposes. We propose to use it for scaling the sEMG signals as a  pre-processing approach. The \textit{$\mu$-law} transformation was performed based on the following formulation
\begin{equation}\label{mu_law}
F(x_t) = \text{sign}(x_t)\frac{\ln{\big(1+ \mu |x_t|\big)}}{\ln{\big(1+ \mu \big)}},
\end{equation}
where $t\geq 1$ is the time index; $x_t$ denotes the input to be scaled, and the parameter $\mu$ defines the new range. Here $\mu = 2,048$ was utilized, i.e., the scaled data points were distributed between $0$ and $2,048$. Afterwards, we fed the scaled sEMG signals to $\minmax$ normalization. We empirically observed that the normalization of the scaled sEMG signals is significantly better than non-scaled sEMG signals.

This completes a brief introduction of the utilized dataset and the introduced pre-processing step. Next, we develop the proposed Meta Learning-based $\EM$ framework.
\section{The $\EM$ Framework}\label{sec:model}
\begin{figure*}[t!]
\centering
\includegraphics[scale=0.92]{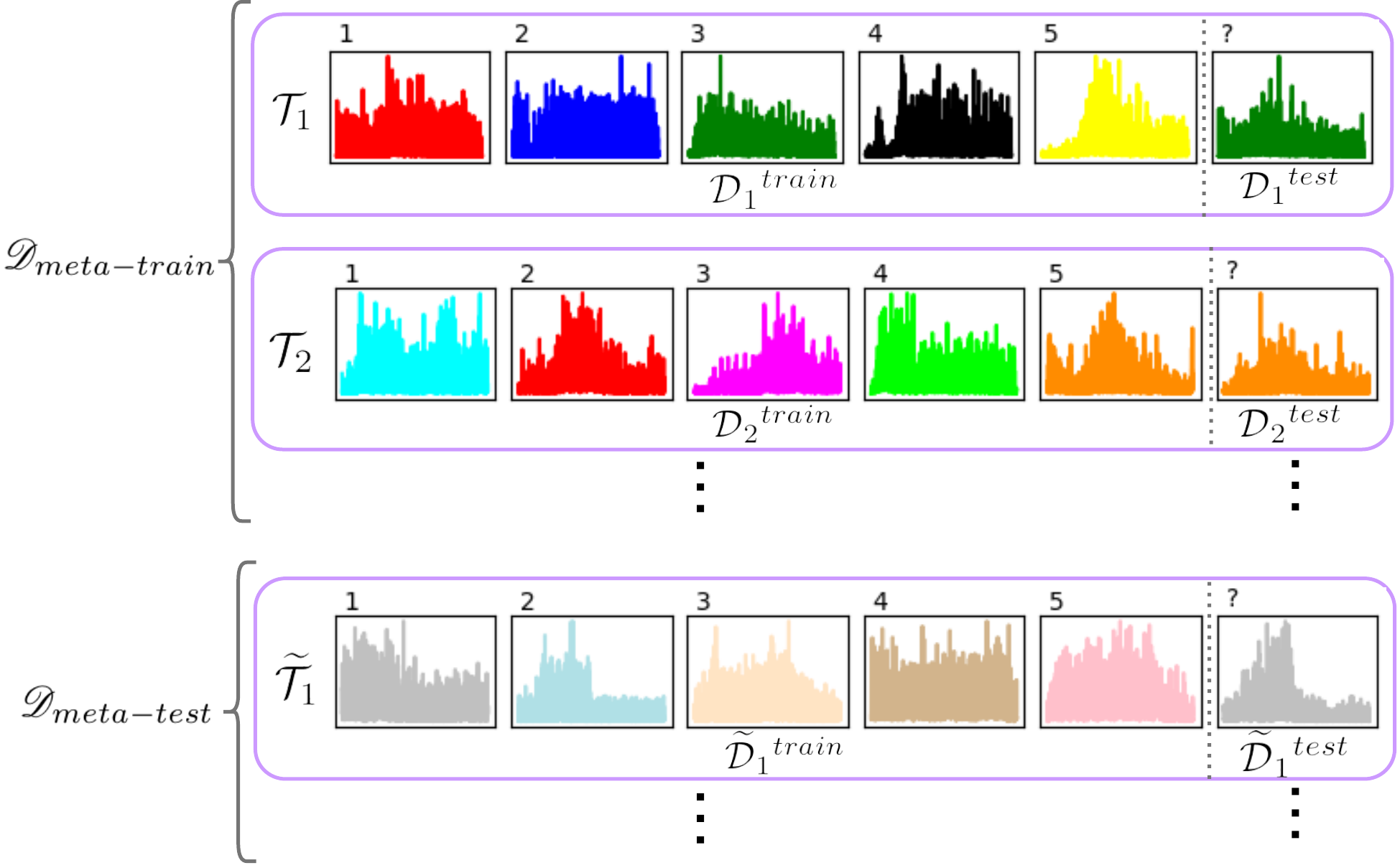}
\vspace{-.15in}
\caption{\small $\mathrm{5}$-way $\mathrm{1}$-shot classification. Each task $\mathcal{T}$, represented in a purple box, is associated with a training set $\mathcal{D}^{train}$ and a prediction set $\mathcal{D}^{test}$. Here, for constructing $\mathcal{D}^{train}$, first, $\mathrm{5}$ classes are sampled from the $\mathscr{D}_{meta-train}$, and then one example from each of these 5 classes (each corresponding with a label $1$-$5$) are sampled. $\mathcal{D}^{test}$ consists of $1$ example sampled from one of those $\mathrm{5}$ classes. The $\mathscr{D}_{meta-test}$ is represented in the same approach, covering a different set of datasets which do not include any classes presented in any of the datasets in $\mathscr{D}_{meta-train}$. Moreover, $\mathscr{D}_{meta-val}$ is defined in the same way to determine the hyper-parameters of the model.}\label{5way-1shot}
\end{figure*}
Meta-learning can be formalized as a sequence-to-sequence learning problem. The bottleneck is in the meta-learner's ability to internalize and refer to experience. To address this shortcoming for the gesture recognition task based on sparse multichannel sEMG, inspired by~\cite{snail}, we proposed a class of model architectures by combining temporal convolutions with attention mechanisms to enable the meta-learner to aggregate contextual information from experience. This integrated architecture allows the meta-learner to pinpoint specific pieces of information within its available set of inputs. Our main goal is to construct and train a hand gesture recognition model that can achieve rapid adaptation. Next, we first elaborate on the meta-learning concept.

\subsection{The Meta-Learning Problem}\label{A}
A supervised learning task starts with a given dataset $\mathcal{D} = \{(\x\i,\mathrm{y_i})\}_{i=1}^{\ND}$, consisting of $\ND$ observations, where the $i^{\text{th}}$ observation is denoted by $\x\i$, for ($1 \leq i \leq N$), with its associated label denoted by $\mathrm{y_i}$. The main objective is to learn a (possibly non-linear) function $f(\cdot)$ defined based on its underlying parameters $\theta$ that maps each observation $\x\i$ to its corresponding label, $\mathrm{y_i} = f(\x\i; \theta)$. In a supervised learning approach, the dataset is divided into: (a) The training data $\mathcal{D}_{train}$ used for learning the parameters $\theta$ of the model; (b) The validation data $\mathcal{D}_{val}$ utilized for tuning the hyper-parameters of the model, and; (c) The test data $\mathcal{D}_{test}$ for model evaluation.

In this context, we focused on meta-supervised learning, where the goal is generalization across tasks rather than across data points. Therefore, instead of using the aforementioned conventional data subsets (Items (a)-(c) above), we have a meta-set denoted by $\mathscr{D}$, which in turn splits into meta-train $\mathscr{D}_{meta-train}$, meta-validation $\mathscr{D}_{meta-val}$, and meta-test $\mathscr{D}_{meta-test}$ sub-datasets. Furthermore, one needs to construct different tasks (as shown in Fig.~\ref{5way-1shot}) within each meta-dataset. Task $\mathcal{T}\j \in \mathscr{D}$ is episodic and is defined by two components, a training set $\mathcal{D}{_j}^{train}$ for learning and a testing set $\mathcal{D}{_j}^{test}$ for evaluation, i.e., $\mathcal{T}_j = (\mathcal{D}{_j}^{train}, \mathcal{D}{_j}^{test})$.

Within the context of meta-learning, our focus is specifically on few-shot learning (typically referred to as $k$-shot learning with $k$ being a small integer), which is briefly described next. In a $\mathit{N}$-way $\mathit{k}$-shot classification, our goal is training on $\mathscr{D}_{meta-train}$, where the input is the training set $\mathcal{D}\j^{train}$ and, a test instance \textbf{x$\j^{test}$} $\in \mathcal{D}{_j}^{test}$. To be more precise, $\mathcal{D}{_j}^{train} = \{(\textbf{x$\i$},\mathrm{y_i})\}_{i=1}^{\mathit{k}\times\mathit{N}} $, where $\mathit{N}$ classes are sampled from the meta-train set, and then $\mathit{k}$ examples are sampled from each of these classes. To make predictions about a new test data point, \textbf{x$\j^{test}$} $\in \mathcal{D}{_j}^{test}$, we produce a mapping function $f(\cdot)$ that takes as input $\mathcal{D}{_j}^{train}$ and \textbf{x$\j^{test}$} to produce the label $\mathrm{\hat{y}\j^{test}} = f(\mathcal{D}^{train}, \textbf{x$\j^{test}$}; \theta)$. Hyper-parameter selection is performed by using $\mathscr{D}_{meta-val}$. Generalization performance of the meta-learner is then evaluated on the $\mathscr{D}_{meta-test}$~\cite{hugo}.

Fig.~\ref{5way-1shot} shows a $N=5$-way $k=1$-shot classification task,
where inside each purple box is a separate dataset $\mathcal{T}_j$ consisting of the training set $\mathcal{D}{_j}^{train}$ (on the Left-Hand Side (LHS) of the dashed line) and the $\mathcal{D}{_j}^{test}$ (on the Right-Hand Side (RHS) of the dashed line). In the illustrative example of Fig.~\ref{5way-1shot}, we are considering a $5$-way $1$-shot  classification task where for each dataset, we have one example from each of the $5$ classes (each given a label $1$ to $5$) in the training set and $1$ example for evaluation from the test set of that specific task.

\subsection{Description of the $\EM$ Model}\label{B}
In few-shot classification, the goal is to reduce the prediction error on data samples with unknown labels given a small training set.  Inspired by~\cite{snail}, the proposed $\EM$ network receives as input a sequence of example-label pairs $\mathcal{D}{_j}^{train} = \{(\textbf{x${_i}$},\mathrm{y_i})\}_{i=1}^{\mathit{k}\times\mathit{N}}$, followed by $\mathcal{D}{_j}^{test}$, which consists of an unlabelled example. The meta-learning model predicts the label of the final example based on the previous labels that it has seen. During the training phase, first, we sample $\mathit{N}$ classes, with $\mathit{k}$ examples per $\mathcal{D}{_j}^{train}$ (in terms of our running illustrative example, for each task, we have $k=1$ sample from each of the underlying $N=5$ classes).
\begin{figure}[t!]
	\centering
	\includegraphics[scale=1.2]{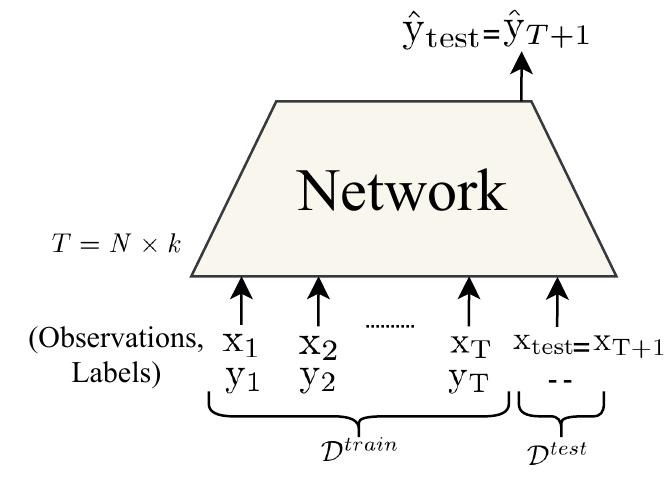}
	\vspace{-.15in}
	\caption{\small For each task $\mathcal{T}_j$, the set of observations and labels are concatenated together and sequentially fed to the model. The final example is concatenated with a null label instead of True label. The network is supposed to predict the missing label of final example based the previous labels that it has seen. In  $\mathit{N}$-way $\mathit{k}$-shot classification,  $\mathit{N}$ shows the number of classes which are sampled from whole set of labels, and $\mathit{k}$ shows the examples that are sampled from each of those $\mathit{N}$ classes. \label{schematic}}
\end{figure}
For constructing the $\mathcal{D}{_j}^{test}$, we sample an extra example from one of those selected classes. Afterwards, each set of the observations and labels are concatenated together (the final example is concatenated with a null label instead of the ground truth label as it is used for evaluation purposes), and then all $(\mathit{N}\times\mathit{k}+1)$ are sequentially fed to the network. Finally, the loss  $\mathcal{L}_j$ is computed between the predicted and ground truth label of the $(\mathit{N}\times\mathit{k}+1)^{th}$ example. During such a training mechanism, the network learns how to encode the first $\mathit{N}\times\mathit{k}$ examples to make a prediction about the final example~\cite{snail}. The training procedure is described in Algorithm~\ref{training} and the schematic of the model is shown in Fig.~\ref{schematic} (further information is available at the link provided in Reference~\cite{Elaheh:Code}).

\begin{algorithm}[t!]
\caption{\textproc{The training procedure}}
\label{training}
\begin{algorithmic}[1]
\Input $\mathscr{D}_{meta-train}$, and; mapping function $f\,(\cdot)$ with parameters $\theta$.
\item[Require.] $\mathit{p}\,(\mathcal{T})$: distribution over tasks
\While{not done}
\State Sample batch of tasks $\mathcal{T}_j\sim\mathit{p}\,(\mathcal{T})$
\ForAll{$\mathcal{T}_j$}
\State Split $\mathcal{T}_j$ into $\mathcal{D}{_j}^{train}$ and $\mathcal{D}{_j}^{test}$
\State Predict the missing label of final example of $\mathcal{T}_j$: $\mathrm{\hat{y}_{test}} = f\,(\mathcal{D}{_j}^{train}, \mathcal{D}{_j}^{test}; \theta)$
\EndFor
\State Update $\theta$ using $\Sigma_{\mathcal{T}_j\sim\mathit{p}\,(\mathcal{T})}\mathcal{L}_{\mathcal{T}_j}\,(\hat{y}_{test}, {y}_{test})$
\EndWhile
\end{algorithmic}
\end{algorithm}
%
\begin{figure}[t!]
\centering
\includegraphics[scale=.85]{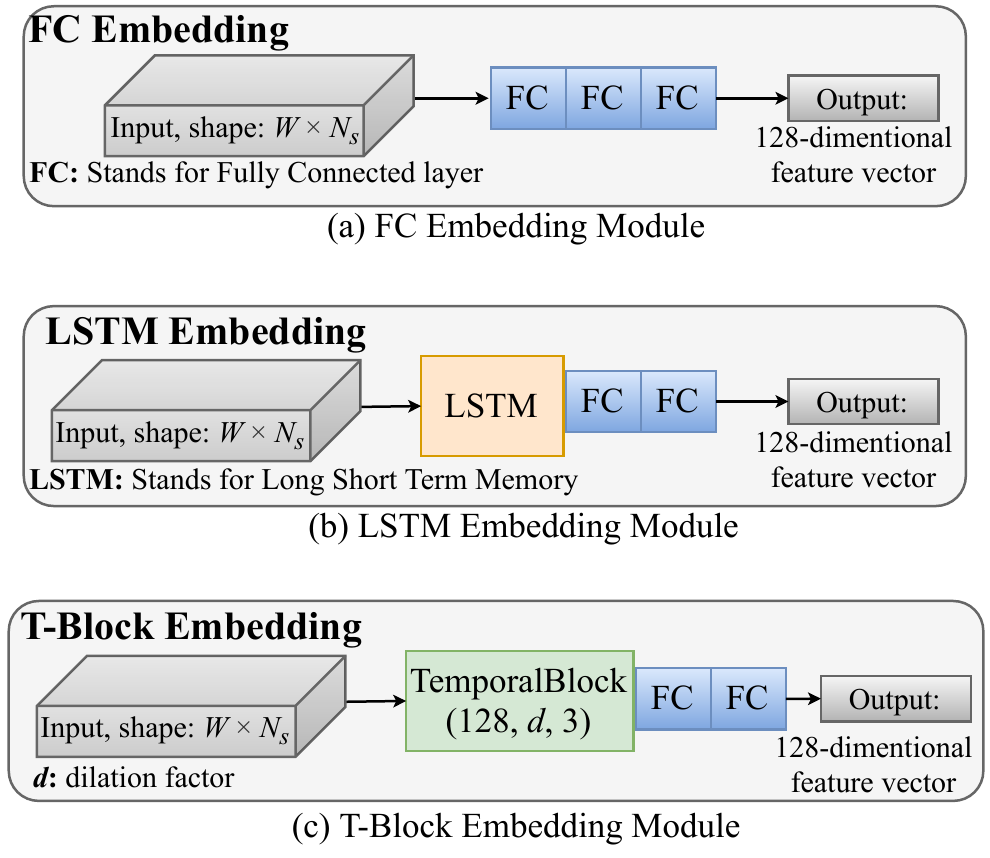}
\vspace{-.09in}
\caption{\small \textbf{The Embedding Module}, which converts an input with size $(W\times\NS$), $W$ stands the sequence length and $\NS$ shows the number of input features, to a $\mathrm{128}$-dimensional feature vector. (a) \textbf{FC Embedding Module}, which uses three FC layers to outputs a $\mathrm{128}$-dimensional feature vector. (b) \textbf{LSTM Embedding Module}, which adopts a LSTM layer followed by two FC layers. (c) \textbf{T-Block Embedding Module}, which consists of a Temporal Block with number of filters $\mathit{f}=128$, kernel size $\mathit{k}=3$, and dilation factor $\mathit{d}$, followed by two FC layers.\label{Embedding}}
\end{figure}
%

\subsection{The Building Modules of the $\EM$ Framework} \label{The Building Modules}
After completion of the pre-processing step, sEMG signals acquired from $\NS$ number of sensors are segmented by a window of length of $W = 200$ ms selected to satisfy the acceptable delay time~\cite{24}, i.e., the window length $W$ is required to be under $300$ ms. Finally, sliding window with steps of $50$ ms is considered for segmentation of the sEMG signals.

%
\begin{figure*}[t!]
\centering
\includegraphics[scale=.7]{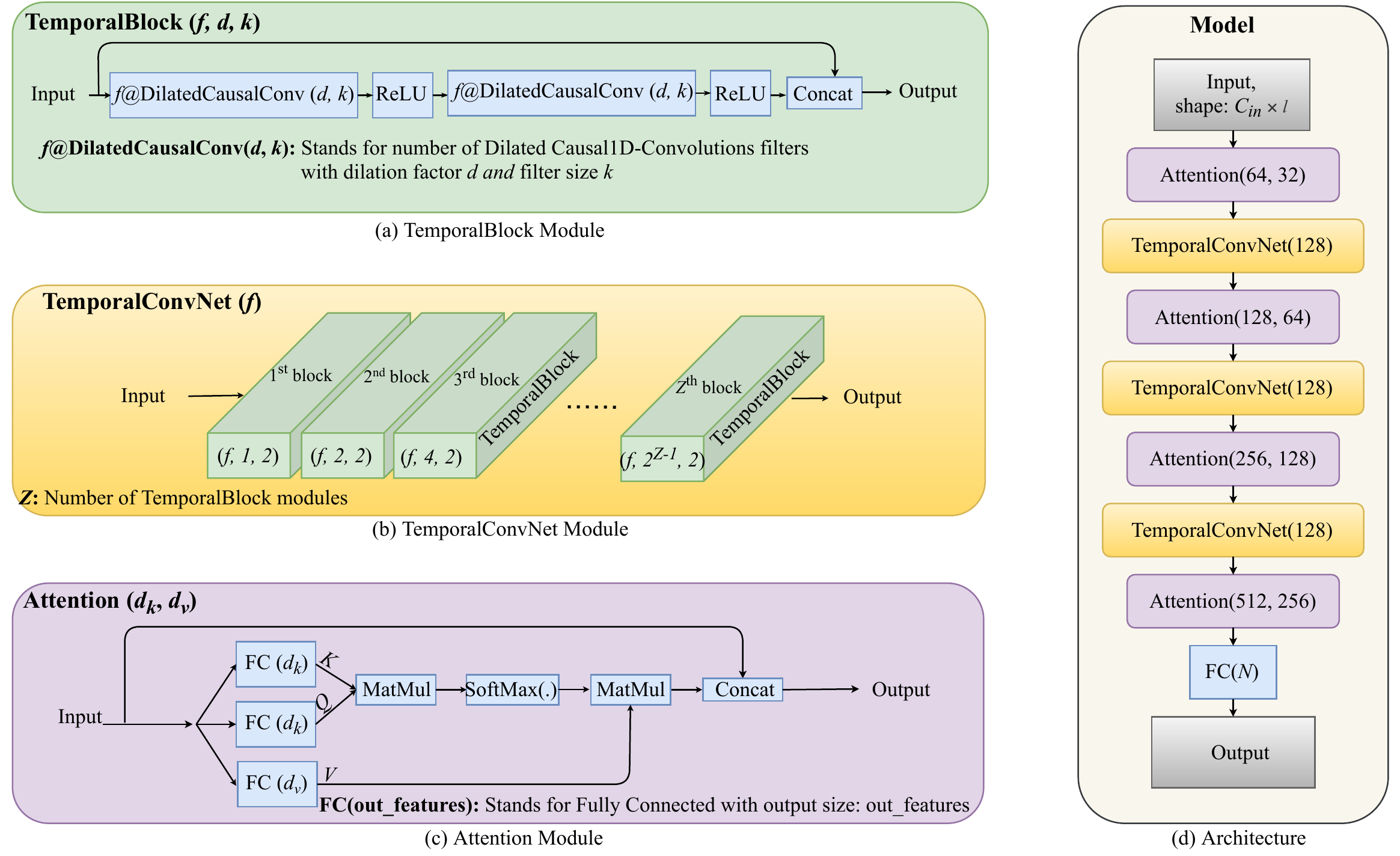}
\vspace{-.1in}
\caption{\small (a) \textbf{The TemporalBlock Module}, which consists of $\mathit{f}$ Dilated Causal 1D-Convolutions with dilation factor $\mathit{d}$ and kernel size $\mathit{k}$. This module converts an input with $\mathit{C}_{in}$ features to an output with $\mathit{C}_{out}$ features. The sequence length of the input $\mathit{l}$ is equal to $(\mathit{N}\times\mathit{k}+1)$, which $\mathit{N}$ shows the number of classes and $\mathit{k}$ denotes the number of examples of each class. (b) \textbf{The TemporalConvNet Module}, which consists of a series of TemporalBlock modules (green ones). The kernel size of each TemporalBlock Module $\mathit{k}$ is equal to $\mathrm{2}$; however, their dilation factor $\mathit{d}$ increases exponentially. (c) \textbf{The Attention Module}, which consists of three FC layers with output size $\mathit{d}_{k}$, $\mathit{d}_{k}$, and $\mathit{d}_{v}$, respectively, to produce matrix $\mathit{Q}$, $\mathit{K}$, and $\mathit{V}.$ (d) \textbf{The Architecture}, consisting of three TemporalConvNet modules (yellow ones), and four Attention modules (purple ones). Here, $\mathrm{128}$ denotes the number of filters $\mathit{f}$ in Dilated 1D-Convolutions. The architecture is supposed to predict the missing label of the  $(\mathit{N}\times\mathit{k} + 1)^{th}$ example in each task $\mathcal{T}_{j}$.\label{model}}
\end{figure*}
%
\subsubsection{\textbf{The Embedding Module}}
To develop the $\EM$  for few-shot learning, we aimed to first extract a $\mathrm{128}$-dimensional feature vector from each observation with size of $(W\times \NS)$, where  $W$ stands for the sequence length and $\NS$ shows the number of input features, e.g., in the experiments $\NS$  is equal to $12$ as twelve sensing channels are available. The ``Embedding Module'' is, therefore, used to extract a $\mathrm{128}$-dimensional feature vector, which is then provided as input to the proceeding modules within the proposed architecture.

Adopting a proper Embedding Module has a significant effect on the results. For validating our claim, therefore, we utilized four different Embedding Modules:
\begin{itemize}
\item[(i)] The first Embedding Module, referred to as the \textit{FC Embedding}, consists of three Fully Connected (FC) layers to output a $\mathrm{128}$-dimensional feature vector from each observation. The first FC layer in the FC Embedding Module is used to increase the input dimensional to $(W\times\mathrm{128})$. Subsequently, the second (which is followed by ReLU activation function) and third FC layers with output size of $\mathrm{100}$ and $\mathrm{1}$, respectively, are adopted to reduce the sequence length of each observation to $(\mathrm{1}\times\mathrm{128})$ (Fig.~\ref{Embedding}(a));
\item[(ii)] \textit{LSTM Embedding}: Fig.~\ref{Embedding}(b) illustrates the second Embedding Module, referred to as the \textit{LSTM Embedding}, which utilizes a Long Short-Term Memory (LSTM) layer as its first block followed by two FC layers.  The LSTM layer takes the observation with input size $\mathrm{12}$ and converts it to an output with $\mathrm{128}$ features. Then, the two FC layers are adopted to reduce the observation's sequence length to $1$;
\item[(iii)] \textit{T-Block Embedding I}: This third Embedding Module utilizes the \textit{TemporalBlock Module} (which will be described in next sub-section) consisting of $\mathit{f}=128$ 1D-Convolutions with kernel size $\mathit{k}=3$, and dilation factor $\mathit{d}=1$ as its first block. The \textit{TemporalBlock Module} is followed by two FC layers to decrease the input's sequence length to $1$ as shown in Fig.~\ref{Embedding}(c), and;
\item[(iv)] \textit{T-Block Embedding II}: This embedding is similar in nature to the one described above in Item (iii), however, here the goal is to examine the effect of increasing the size of the receptive field. As such, the fourth Embedding Module utilizes two \textit{TemporalBlock Modules} with $\textit{d}=1$ and $\textit{d}=2$. It is noteworthy to mention that the first FC layer in both LSTM and T-Block Embedding modules are followed by ReLU activation function.
\end{itemize}
%

\subsubsection{\textbf{The TemporalBlock Module}}
Inspired by~\cite{JMRR_Elahe, Globalsip_Elahe, snail, 10, 11}, the proposed $\EM$  few-shot learning architecture utilizes \textit{Dilated Causal 1D-Convolutions} over the temporal dimension. The proposed architecture, therefore, provides several advantages over RNNs such as low memory requirement and faster training. In addition, and unlike conventional CNNs, by incorporation of dilated causal convolutions, we increased the receptive field of the network and as such benefit from the time-series nature of the input.

As shown in Fig.~\ref{model}(a), each \textit{TemporalBlock} consists of two dilated causal 1D-convolutions, each with dilation factor $\mathit{d}$, filter size $\mathit{k}$, and $\mathit{f}$ number of filters. To learn the complex structure of the underlying data, each Dilated Causal 1D-Convolutions is followed by a ReLU activation function. Finally, by concatenating the results and the input, the training speed can be considerably improved. This module takes an input with size $(\mathit{C}_{in}\times\mathit{l})$ and output a tensor with size $(\mathit{C}_{out}\times\mathit{l})$. Here, $\mathit{l}$ denotes the sequence length and is equal to $(\mathit{N}\times\mathit{k} + 1)$, which $\mathit{N}$ shows the number of class samples from the whole set of labels, and $\mathit{k}$ shows the number of examples per each class.
\subsubsection{\textbf{The TemporalConvNet Module}}
The benefit that comes with the designed ``TemporalConvNet'' module is that its training procedure is much faster and efficient  compared to LSTM or Gated Recurrent Unit (GRU) architectures. In other words, through this approach one complete sequence can be processed through only one forward pass, while in RNN-based models this, typically, needs several passes due to temporally linear hidden state dependency.
The TemporalConvNet module consists of a series of TemporalBlock modules with exponentially growing dilation factors $\mathit{d}$. More specifically, as shown in Fig.~\ref{model}(b),  for an input with sequence length $\mathit{l}=(\mathit{N}\times\mathit{k}+1)$, the TemporalConvNet consists of $Z=\left \lceil{\log_2 l}\right \rceil$ number of TemporalBlock modules. The dilation factors $\mathit{d}$ for the TemporalBlock modules are equal to $\mathrm{[1, 2, 4, ..., 2^{\mathit{Z}-1}]}$, respectively.

\subsubsection{\textbf{The Attention Module}}
The final constituent module within the proposed $\EM$ architecture is referred to as the ``Attention Module," included with the objective of pinpointing a specific type of information within the available (possibly significantly large) context~\cite{Vaswani}. Attention mechanism has been recently utilized~\cite{Josephs:2020} within the context of  sEMG-based hand gesture recognition, where the experiments showed attention's capability to learn a time-domain representation of multichannel sEMG data. By integrating the TemporalConvNet, described above, and the Attention Module, essentially we provided the $\EM$ architecture with the capability to access the past experience without any limitations on the size of experience that can be used effectively. Furthermore, in the $\EM$ framework we used the Attention Module at different stages to provide the model with the ability to learn  how to identify and select pieces of useful information and its appropriate representation from its  experience.

As shown in Fig.\ref{model}(c), to get queries, keys, and values, three linear transformations are applied to the input. The attention mechanism then compares queries to each of the key values with a dot-product, scaled by $\sqrt{\mathit{d_k}}$, which results compatibility scores.
To obtain attention distribution over the values, softmax function is applied to the scores. Then, we computed the weighted average of the values, weighted by the attention distribution.
In practice, the keys, values, and queries are packed together into matrices  $\bm{K}$, $\bm{V}$, and $\bm{Q}$, respectively. The matrix of outputs is obtained as follows:
\begin{equation}\label{attention}
\text{Attention}(\bm{Q}, \bm{K}, \bm{V}) = \text{softmax}(\frac{\bm{Q}\bm{K}^T}{\sqrt{d_k}})\bm{V},
\end{equation}
where $\mathit{d_k}$ stands for length of the key vector in matrix $\bm{K}$. Then, the results and inputs are concatenated together. This completes description of the modules incorporated to construct the proposed $\EM$ framework. Next, we present its overall architecture.

\subsection{The Architecture}
The overall  structure of the proposed $\EM$ architecture consists of four Attention modules, where the first three ones are followed by a TemporalConvNet module. The final Attention module is followed by a FC layer to produce the label of the final example in each task $\mathcal{T}_{j}$. More specifically, after feeding each observation with size $\textit{W}\times\textit{N}_S$ to an Embedding Module, we obtained a $\mathrm{128}$-dimensional feature vector (Fig.~\ref{Embedding}). Then, for constructing each task $\mathcal{T}_{j}$ with sequence length $\mathit{l}$ (Fig.~\ref{schematic}), the set of observations (each observation is converted to a $\mathrm{128}$-dimensional feature vector) and labels are concatenated. The final observation in the sequence is concatenated with a null label instead of a True label. The network is supposed to predict the missing label of the final example based on the previous labels that it has seen. In summary, to perform the hand gesture recognition task, the $\EM$ framework is constructed based on different modules as shown in Fig.~\ref{model}(d).

\section{Experiments and Results}\label{sec:results}
In this section, we describe a comprehensive set of experiments to analyse and evaluate the proposed $\EM$ framework. It is worth mentioning that in few-shot classification, we would like to classify inputs in $\mathit{N}$ classes when we have just $\mathit{k}$ examples per class. To evaluate the proposed architecture for $\mathit{N}$-way $\mathit{k}$-shot classification, we randomly sampled $\mathit{N}$ classes from the overall classes, and then sampled $\mathit{k}$ examples from each class. Then, we fed the $(\mathit{N}\times\mathit{k})$ observation-label pairs to the model followed by a new unlabelled example sampled from one of the $\mathit{N}$ classes. The objective of the $\EM$ model is to predict the missing label of the $(\mathit{N}\times\mathit{k} +1)^{th}$ in the sequence.

In the following, we present three evaluation scenarios. In all experiments, Adam optimizer was used for training purposes with learning rate of $0.0001$. Different models were trained with a mini-batch size of $64$ except in $10$-way $5$-shot classification where mini-batch size of $32$ was used. For measuring the classification performance, the loss $\mathcal{L}_j$ was computed between the predicted and ground truth label of $(\mathit{N}\times\mathit{k}+1)^{th}$ example in each task $\mathcal{T}_{j}$. The average loss was computed using Cross-entropy loss. Finally, the average accuracy is reported on the $(\mathit{N}\times\mathit{k} +1)^{th}$ example.

\begin{table}[t!]
	\centering
	\renewcommand\arraystretch{2}
	\caption{\small Experiment 1: $5$-way, $1$-shot, $5$-shot, and $10$-shot classification accuracies on \textit{new repetitions with few-shot observation}. The classification on new repetitions with few-shot observation are performed by using Meta-supervised Learning approach. This table also shows a comparison between our methodology (Meta-supervised) learning and previous works where Supervised learning methodology is used.\label{table3}}
	\resizebox{\columnwidth}{!}
	{\begin{tabular}{  c | c c c c}
			\hline
			\hline
			\multicolumn{1}{c}{\multirow{6}[2]{*}{\rotatebox[origin=c]{90}{\textbf{Meta-Supervised Learning}}}} \newline \multirow{6}[2]{*}{\rotatebox[origin=c]{90}{\textbf{Proposed Method}}}
			& \multicolumn{1}{|c|}{\multirow{1}[5]{*}{\textbf{The Embedding}}}
			& \multicolumn{3}{c}{\textbf{5-way Accuracy}} \\
			\cline{3-5}
			&
			\multicolumn{1}{|c|}{\multirow{1}[-7]{*}{\textbf{Module}}}
			& \multicolumn{1}{|c}{\textbf{1-shot}}
			& \textbf{5-shot} & \textbf{10-shot}
			\\
			\cline{2-5}
			&
			\multicolumn{1}{|c|}{\textbf{FC Embedding}} & 72.59$\%$ & 85.13$\%$ & 89.26$\%$
			\\
			&
			\multicolumn{1}{|c|}{\textbf{LSTM Embedding}} & \textbf{75.03}$\%$ & 84.06$\%$ & 88.45$\%$
			\\
			&
			\multicolumn{1}{|c|}{\textbf{T-Block Embedding I}} & 73.46$\%$ & \textbf{85.94}$\%$ & 89.40$\%$
			\\
			&
			\multicolumn{1}{|c|}{\textbf{T-Block Embedding II}} & 74.89$\%$ & 85.88$\%$  & \textbf{89.70}$\%$
			\\
			\hline
			\multicolumn{1}{c}{\multirow{7}[4]{*}{\rotatebox[origin=c]{90}{\textbf{Supervised Learning}}}} \newline \multirow{6}[10]{*}{\rotatebox[origin=c]{90}{\textbf{Previous Works}}}
			& \multicolumn{1}{|c|}{\textbf{Previous Works}}
			& \multicolumn{2}{|c}{\multirow{2}[-5]{-12cm}{\textbf{Accuracy}}}
			\\
			\cline{2-5}
			&
			\multicolumn{1}{|c|}{Wei \textit{et al.}~\cite{WeiNet}} & \multicolumn{2}{|c}{\multirow{2}[-5]{-12cm}{\textbf{83.70$\%$}}}
			\\
			&
			\multicolumn{1}{|c|}{Hu et al.~\cite{YuNet}} & \multicolumn{2}{|c}{\multirow{2}[-5]{-12cm}{82.20$\%$}}
			\\
			&
			\multicolumn{1}{|c|}{Ding \textit{et al.}~\cite{DingNet}}& \multicolumn{2}{|c}{\multirow{2}[-5]{-12cm}{78.86$\%$}}
			\\
			&
			\multicolumn{1}{|c|}{Zhai \textit{et al.}~\cite{ZhaiNet}} & \multicolumn{2}{|c}{\multirow{2}[-5]{-12cm}{78.71$\%$}}
			\\
			&
			\multicolumn{1}{|c|}{Geng \textit{et al.}~\cite{GengNet}} & \multicolumn{2}{|c}{\multirow{2}[-5]{-12cm}{77.80$\%$}}
			\\
			&
			\multicolumn{1}{|c|}{Atzori \textit{et al.}~\cite{AtzoriNet}} & \multicolumn{2}{|c}{\multirow{2}[-5]{-12cm}{75.27$\%$}}
			\\
			
			\hline
			\hline
		\end{tabular}}
	\end{table}

	\vspace{.3in}
	\noindent
	\textbf{\textit{Experiment 1: Classification on New-Repetitions with Few-Shot Observation}}.
	The first experiment shows that our proposed network is applicable when we had new repetitions with few-shot observation on the target. We evaluated our proposed architecture when $\mathscr{D}_{meta-train}$ consisted of the $2/3$ of the gesture trials of each subject (following Reference~\cite{AtzoriNet}, repetitions $1$, $3$, $4$, and $6$ repetitions were used for training purposes), and $\mathscr{D}_{meta-test}$ consisted of the remaining repetitions. Table~\ref{table3} shows our results when using few-shot classification as well as previous works which used supervised learning. From Table~\ref{table3}, it can be observed that the proposed $\EM$ architecture outperformed existing methodologies when evaluated based on the same setting, i.e., $85.94$\% best accuracy with the $\EM$ compared to $83.70$\% best accuracy achieved by the state-of-the-art. Although this improvement is relatively small, the following Experiments 2 and 3 provide further evidence for the superior performance of the proposed approach.
\begin{table}[t!]
\centering
\renewcommand\arraystretch{2}
\caption{\small Experiment 2(a): $5$-way and $10$-way, $1$-shot and $5$-shot classification accuracies based on \textit{new subjects with few-shot observation}. In this experiment, we adopted four different Embedding Modules: (i) FC Embedding; (ii) LSTM Embedding; (iii) T-Block Embedding I, and; (iv) T-Block Embedding~II.}
\label{table1}
\resizebox{\columnwidth}{!}
{\begin{tabular}{  c | c c c c}
\hline
\hline
\multicolumn{1}{c|}{\multirow{2}[2]{*}{\textbf{The Embedding Module}}} & \multicolumn{2}{|c|}{\textbf{5-way Accuracy}}
& \multicolumn{2}{|c}{\textbf{10-way Accuracy}} \\
\cline{2-5}
&  \textbf{1-shot}
&  \multicolumn{1}{c|}{\textbf{5-shot}}
&\textbf{1-shot}
& \textbf{5-shot}
\\
\hline
\textbf{FC Embedding}
& 62.87$\%$
&\multicolumn{1}{c|}{78.90$\%$}
& 43.47$\%$
& 68.59$\%$
\\
\textbf{LSTM Embedding}
& 64.46$\%$
& \multicolumn{1}{c|}{79.82$\%$ }
& 49.58$\%$
& 69.93$\%$
\\
\textbf{T-Block Embedding I}
& \textbf{67.81}$\%$
& \multicolumn{1}{c|}{81.08$\%$ }
& 50.31$\%$
& 69.94$\%$
\\
\textbf{T-Block Embedding II}
& 66.98$\%$
& \multicolumn{1}{c|}{\textbf{81.29}$\%$ }
& \textbf{52.05}$\%$
&  \textbf{70.71}$\%$
\\

\hline
\hline
\end{tabular}}
\end{table}

\vspace{.1in}
\noindent
\textbf{\textit{Experiments 2: Classification on New-Subject with Few-Shot Observation}}. In this scenario, like the previous experiment, the second Ninapro database DB2 was utilized. It consists of $\mathrm{49}$ gestures plus rest from $\mathrm{40}$ intact-limb subjects. In this experiment, to validate our claim that the proposed $\EM$ architecture can classify hand gestures of new subjects just by training with a few examples, we split the DB2 database into  $\mathscr{D}_{meta-train}$,  $\mathscr{D}_{meta-val}$, and  $\mathscr{D}_{meta-test}$ such that the subjects in these meta-sets are completely different (i.e., there is no overlap between the meta-sets). In other words, $\mathscr{D}_{meta-train}$ consists of the first $\mathrm{27}$ subjects, while  $\mathscr{D}_{meta-val}$ includes the sEMG signals from the $\mathrm{28}^{th}$ subject to $\mathrm{32}^{ed}$ subject (5 subjects). Finally, we evaluated our model on the remaining subjects, i.e., $\mathscr{D}_{meta-test}$ consists of the final $\mathrm{8}$ subjects in the DB2 database.

It is noteworthy to mention that the proposed network is trained once and shared across all participants (which is different from previous works that trained the model separately for each participant). For constructing task $\mathcal{T}_{j}$, however, we can feed data in two different approaches:
\begin{itemize}
\item \textit{Experiment 2(a)}: In the first approach,  for constructing $\mathcal{D}{_j}^{train}$ for each task $\mathcal{T}_{j}$, we sampled all of the $\mathit{N}$ classes from a specific user, which was randomly selected from the existing participants. This is the more realistic and practical scenario.
\item \textit{Experiment 2(b)}: In the second approach, for constructing $\mathcal{D}{_j}^{train}$, $\mathit{N}$ classes were sampled from different participants.
\end{itemize}
\begin{table}[t!]
\centering
\renewcommand\arraystretch{2}
\caption{\small Comparison of $5$-way, $1$-shot and $5$-shot classification accuracies between the Experiment 2(a) and 2(b) based on \textit{new subjects with few-shot observation.}}
\label{table_subject}
\resizebox{\columnwidth}{!}
{\begin{tabular}{  c | c c c c}
\hline
\hline
\multicolumn{1}{c|}{\multirow{3}[2]{*}{\textbf{The Embedding Module}}}
& \multicolumn{2}{|c|}{\textbf{Experiment 2(a)}}
& \multicolumn{2}{|c}{\textbf{Experiment 2(b)}}
\\
\cline{2-5}
&
\multicolumn{4}{c}{\multirow{1}[2]{*}{\textbf{5-way Accuracy}}}
\\
&  \textbf{1-shot}
&  \multicolumn{1}{c|}{\textbf{5-shot}}
&\textbf{1-shot}
& \textbf{5-shot}
\\
\hline
\textbf{FC Embedding}
& 62.87$\%$
&\multicolumn{1}{c|}{78.90$\%$}
& 72.69$\%$
& 86.08$\%$
\\
\textbf{LSTM Embedding}
& 64.46$\%$
& \multicolumn{1}{c|}{79.82$\%$ }
& 75.56$\%$
& 89.14$\%$
\\
\textbf{T-Block Embedding I}
& \textbf{67.81}$\%$
& \multicolumn{1}{c|}{81.08$\%$ }
& 75.11$\%$
& 89.66$\%$
\\
\textbf{T-Block Embedding II}
& 66.98$\%$
& \multicolumn{1}{c|}{\textbf{81.29}$\%$ }
& \textbf{77.08}$\%$
&  \textbf{90.47}$\%$
\\

\hline
\hline
\end{tabular}}
\end{table}

Table~\ref{table1} shows few-shot classification accuracies for Experiment 2(a) based on four different embedding modules. The adaptive learning method of the proposed $\EM$ focuses on transfer learning information between a source and a target domain despite the existence of a distribution mismatch between $\mathscr{D}_{meta-train}$ and $\mathscr{D}_{meta-test}$. The results reported in Table~\ref{table1} show that the proposed mechanism achieves acceptable results despite the fact that the sEMG signals are user-dependant. Table~\ref{table_subject} shows a comparison of $5$-way classification accuracies between Experiments 2(a) and 2(b). As was it expected, Experiment 2(b) achieved better results, which is due to the presence of variations among the probability distribution of sEMG signals obtained from different subjects. However, this is not a practical setting as in practice all of the $\mathit{N}$ classes in $\mathcal{D}{_j}^{train}$ comes from the same user (i.e., Experiment 2(a)).  Experiment 2(a) is the more realistic and challenging one. Experiment 2(b) is included for completeness and comparison purposes.

Finally, it is worth noting that adoption of few-shot learning within the $\EM$ framework has resulted in significant reduction in the required training time for users. As explained before, the dataset was collected from $40$ people including $49$ gesture with $6$ repetition of each, where each repetition lasted $5$ seconds. In previous studies, $4$ repetitions, 20 seconds in total, of each user's gestures were considered for the training purpose, and the remaining $2$ repetitions were used for their model evaluation. Commonly, sliding-window with window size of $200$ ms is leveraged for feeding data to the models. However, in our few-shot based model, we used 6 repetitions of $27$ subjects for training, and the model did not see any data of the remaining subjects during the learning procedure. The gained experience during the training is leveraged to tune the model to a new user by seeing a small number of intervals (each of duration $200$ ms). More specificity, for a new user in a $\mathit{N}$-way $\mathit{k}$-shot classification problem, we just used $\mathit{N}\times\mathit{k}$ windows. For example, in $5$-way $1$-shot, we just need $5$ windows ($1$ second in total) to recalibrate the model. Here, unlike previous methods, we do not have to train the model from scratch for a new user, or fine tune model for $4$ repartitions of new user's gestures. We needed only $\mathit{N}\times\mathit{k}$ windows from a new user, to adapt the trained model for this new user. Therefore, by using few-shot learning, the proposed $\EM$ framework has the potential to significantly  reduce the training time.

\begin{table}[t!]
	\centering
	\renewcommand\arraystretch{2}
	\caption{\small Experiment 3: $5$-way, $1$-shot, $5$-shot, and $10$-shot classification accuracies based on \textit{new gesture with few-shot observation}.\label{table2}}
	\resizebox{\columnwidth}{!}
	{\begin{tabular}{  c | c c c}
			\hline
			\hline
			\multicolumn{1}{c|}{\multirow{2}[2]{*}{\textbf{The Embedding Module}}} & \multicolumn{3}{|c}{\textbf{5-way Accuracy}}
			\\
			\cline{2-4}
			&  \textbf{1-shot}
			& \textbf{5-shot}
			& \textbf{10-shot}
			\\
			\hline
			\textbf{FC Embedding}
			& 45.94$\%$
			& 67.20$\%$
			& 79.87$\%$
			\\
			\textbf{LSTM Embedding}
			& 46.05$\%$
			& 71.76$\%$
			& 81.58$\%$
			
			\\
			\textbf{T-Block Embedding I}
			& \textbf{49.78}$\%$
			& 71.57$\%$
			& \textbf{83.41}$\%$
			\\
			\textbf{T-Block Embedding II}
			& 45.48$\%$
			& \textbf{73.36}$\%$
			& --
			\\
			\hline
			\hline
		\end{tabular}}
	\end{table}
\vspace{.1in}
\noindent
\textbf{\textit{Experiment 3: Classification on New-Gestures with few-shot observations}}.
In this scenario, the goal is evaluating the capability of the proposed $\EM$ architecture when the target consists of solely out-of-sample gestures (i.e., new gestures with few-shot observation). Performing well in this task allows the model to evaluate new observations, exactly one per novel hand gesture class.
In this experiment, the Ninapro database DB2 was used. The DB2 dataset includes three sets of exercises denoted by Exercise $B$, $C$, and $D$. Exercise $B$ includes $8$ isometric and isotonic hand configurations and $9$ basic movements of the wrist; Exercise $C$ consists of $23$ grasping and functional movements; and finally, Exercise $D$ consists of $9$ force patterns. For training purposes, $\mathscr{D}_{meta-train}$ consisted of the first $34$ gestures of each user, which is equal to approximately $68\%$ of the total gestures. $\mathscr{D}_{meta-val}$ included $6$ gestures or $12\%$ of the total gestures. The remaining gestures ($9$ gestures), were used in $\mathscr{D}_{meta-test}$ for evaluation purposes. Exercises $B$ and $C$ were, therefore, used for training and validation, and Exercises $D$, with different gestures, were used for test purposes. Table~\ref{table2} shows the efficiency of the proposed model when we had out-of-sample gestures in the target. The model predicted unknown class distributions in scenarios where few examples from the target distribution were available.

\section{Conclusion}  \label{sec:con}
We proposed a novel few-shot learning recognition approach for the task of hand gesture recognition via sEMG signals. The proposed $\EM$ framework could quickly generalize after seeing very few examples from each class. This is achieved by  exploiting the knowledge gathered from previous experiences to accelerate the learning process performed by a new subject. The experience gained over several source subjects is leveraged to reduce the training time of a new target user. In this way the learning process does not start every time from the beginning, and instead refines.
The ability to learn quickly based on a few examples is a key characteristic of the proposed $\EM$ framework that distinguishes this novel architecture from its previous counterparts.
A second contribution of the paper is its capability to address the user-dependent nature of the sEMG signals. The proposed $\EM$ framework transfers  information between a source and a target domain despite the existence of a distribution mismatch among them. This would dramatically reduce the number of required cumbersome training sessions leading to a drastic reduction in functional prosthesis abandonment.

\section*{Acknowledgement}
This work was supported by Borealis AI through the Borealis AI Global Fellowship Award.

\bibliographystyle{IEEEbib}

\begin{thebibliography}{46}

\bibitem{2_Dario}
N. Jiang, S. Dosen, K.R. Muller, D. Farina,
\newblock ``Myoelectric Control of Artificial Limbs- Is There a Need to Change Focus?''
\newblock {\em IEEE Signal Process. Mag.}, vol. 29, pp. 150-152, 2012.

\bibitem{Dario}
D. Farina, R. Merletti, R.M. Enoka,
\newblock ``The Extraction of Neural Strategies from the Surface EMG,''
\newblock {\em J. Appl. Physiol.}, vol. 96, pp. 1486-95, 2004.

\bibitem{LDA-SVM}
D. Esposito, E. Andreozzi, G.D. Gargiulo, A. Fratini, G. D'Addio, G.R. Naik, and P. Bifulco,
\newblock ``A Piezoresistive Array Armband with Reduced Number of Sensors for Hand Gesture Recognition,''
\newblock {\em Frontiers in Neurorobotics}, vol. 13, p. 114, 2020.

\bibitem{SVM}
M. Tavakoli, C. Benussi, P.A. Lopes, L.B. Osorio, and A.T. de Almeida,
\newblock ``Robust Hand Gesture Recognition with a Double Channel Surface EMG Wearable Armband and SVM Classifier,''
\newblock {\em Biomedical Signal Processing and Control}, vol. 46, pp. 121-130, 2018.

\bibitem{LDA}
G.R. Naik, A.H. Al-Timemy, H.T. Nguyen,
\newblock ``Transradial Amputee Gesture Classification using an Optimal Number of sEMG Sensors: an Approach using ICA Clustering,''
\newblock {\em IEEE Trans. Neural Syst. Rehabil. Eng.}, vol. 24, no. 8, pp. 837–846, 2015.

\bibitem{DB5}
S. Pizzolato, L. Tagliapietra, M. Cognolato, M. Reggiani, H. Muller, and M. Atzori,
\newblock ``Comparison of Six Electromyography Acquisition Setups on Hand Movement Classification Tasks,''
\newblock {\em PLoS ONE}, vol. 12, no. 10, pp. 1-7, 2017.

\bibitem{19-Patrick3}
C. Castellini, P. Artemiadis, M. Wininger, A. Ajoudani, M. Alimusaj, A. Bicchi, B. Caputo, W. Craelius, S. Dosen, K. Englehart, and D. Farina
\newblock ``Proceedings of the First Workshop on Peripheral Machine Interfaces: Going Beyond Traditional Surface Electromyography,''
\newblock {\em Frontiers in neurorobotics}, 8, p. 22, 2014.

\bibitem{3_Dario}
D. Farina, N. Jiang, H. Rehbaum, A. Holobar, B. Graimann, H, Dietl, and O. C. Aszmann,
\newblock ``The Extraction of Neural Information from the Surface EMG for the Control of Upper-limb Prostheses: Emerging Avenues and Challenges.,''
\newblock {\em IEEE Trans. Neural Syst. Rehabil. Eng.},  vol. 22, no.4, pp. 797-809, 2014.

\bibitem{FD-TBME}
A. K. Clarke \textit {et al.},
\newblock ``Deep Learning for Robust Decomposition of High-Density Surface EMG Signals,"
\newblock {\em IEEE Transactions on Biomedical Engineering}, 2020, In Press.

\bibitem{JMRR_Elahe}
E. Rahimian, S. Zabihi, S. F. Atashzar, A. Asif, and A. Mohammadi,
\newblock ``Surface EMG-Based Hand Gesture Recognition via Hybrid and Dilated Deep Neural Network Architectures for Neurorobotic Prostheses,''
\newblock {\em Journal of Medical Robotics Research}, 2020, pp. 1-12.

\bibitem{Icassp_Elahe}
E. Rahimian, S. Zabihi, F. Atashzar, A. Asif, A. Mohammadi,
\newblock ``XceptionTime: Independent Time-Window XceptionTime Architecture for Hand Gesture Classification,''
\newblock {\em International Conference on Acoustics, Speech, and Signal Processing (ICASSP)}, 2020.

\bibitem{Globalsip_Elahe}
E. Rahimian, S. Zabihi, S. F. Atashzar, A. Asif, and A. Mohammadi,
\newblock ``Semg-based Hand Gesture Recognition via Dilated Convolutional Neural Networks,''
\newblock {\em Global Conference on Signal and Information Processing, GlobalSIP}, 2019.

\bibitem{Josephs:2020}
D. Josephs, C. Drake, A. Heroy, and J. Santerre,
\newblock ``sEMG Gesture Recognition with a Simple Model of Attention,''
\newblock {\em arXiv preprint arXiv:2006.03645}, 2020.

\bibitem{sensor2020}
L. Chen, J. Fu, Y. Wu, H. Li, and B. Zheng,
\newblock ``Hand Gesture Recognition using Compact CNN via Surface Electromyography Signals,''
\newblock {\em Sensors}, vol. 20, no.3,  p. 672,  2020.

\bibitem{Peng:2020}
Y. Peng, H. Tao, W. Li, H. Yuan and T. Li,
\newblock ``Dynamic Gesture Recognition based on Feature Fusion Network and Variant ConvLSTM,"
\newblock {\em IET Image Processing}, vol. 14, no. 11, pp. 2480-2486, 18 9 2020.

\bibitem{pattern_letter2019}
W. Wei, Y. Wong, Y. Du, Y. Hu, M. Kankanhalli, and W. Geng,
\newblock ``A Multi-stream Convolutional Neural Network for sEMG-based Gesture Recognition in Muscle-computer Interface,''
\newblock {\em Pattern Recognition Letters}, 119, pp. 131-138,  2019.

\bibitem{WeiNet}
W. Wei, Q. Dai, Y. Wong, Y. Hu, M. Kankanhalli, and W. Geng,
\newblock ``Surface Electromyography-based Gesture Recognition by Multi-view Deep Learning,''
\newblock {\em IEEE Trans. Biomed. Eng.}, vol. 66, no. 10, pp. 2964-2973, 2019.

\bibitem{YuNet}
Y. Hu, Y. Wong, W. Wei, Y. Du, M. Kankanhalli, and W. Geng,
\newblock ``A Novel Attention-based Hybrid CNN-RNN Architecture for sEMG-based Gesture Recognition,''
\newblock {\em PloS one 13}, no. 10,  2018.

\bibitem{DingNet}
Z. Ding, C. Yang, Z. Tian, C. Yi, Y. Fu, and F. Jiang,
\newblock ``sEMG-based Gesture Recognition with Convolution Neural Networks,''
\newblock {\em Sustainability 10}, no. 6, p. 1865,  2018.

\bibitem{ZhaiNet}
X. Zhai, B. Jelfs, R. H. Chan, and C. Tin,
\newblock ``Self-recalibrating Surface EMG Pattern Recognition for Neuroprosthesis Control based on Convolutional Neural Network,''
\newblock {\em Frontiers in neuroscience}, 11, p.379,  2017.

\bibitem{GengNet}
W. Geng, Y. Du, W. Jin, W. Wei, Y. Hu, and J. Li,
\newblock ``Gesture Recognition by Instantaneous Surface EMG Images,''
\newblock {\em Scientific reports}, 6, p. 36571,  2016.

\bibitem{AtzoriNet}
M. Atzori, M. Cognolato, and H. Müller,
\newblock ``Deep Learning with Convolutional Neural Networks Applied to Electromyography Data: A Resource for the Classification of Movements for Prosthetic Hands,''
\newblock {\em Frontiers in neurorobotics 10}, p.9, 2016.

\bibitem{1_Ninapro}
M. Atzori, A. Gijsberts, C. Castellini, B. Caputo, A.G.M Hager, S. Elsig, G. Giatsidis, F. Bassetto, and H. Müller,
\newblock ``Electromyography Data for Non-Invasive Naturally-Controlled Robotic Hand Prostheses,''
\newblock {\em Scientific data 1}, no. 1, pp. 1-13,  2014.

\bibitem{2_Ninapro}
A. Gijsberts, M. Atzori, C. Castellini, H. Müller, and B. Caputo,
\newblock ``Movement Error Rate for Evaluation of Machine Learning Methods for sEMG-based Hand Movement Classification,''
\newblock {\em IEEE Trans. Neural Syst. Rehabil. Eng.}, vol. 22, no. 4, pp. 735-744,  2014.

\bibitem{3_Ninapro}
M. Atzori, A. Gijsberts, I. Kuzborskij, S. Heynen, A.G.M Hager, O. Deriaz, C. Castellini, H. Mller, and B. Caputo,
\newblock ``A Benchmark Database for Myoelectric Movement Classification,''
\newblock {\em IEEE Trans. Neural Syst. Rehabil. Eng.}, 2013.

\bibitem{snail}
N. Mishra, M. Rohaninejad, X. Chen, and P. Abbeel,
\newblock ``A Simple Neural Attentive Meta-Learner,''
\newblock {\em arXiv preprint arXiv:1707.03141}, 2017.

\bibitem{hugo}
S. Ravi, and H. Larochelle,
\newblock ``Optimization as a Model for Few-shot Learning,''
\newblock {\em  International Conference on Learning Representations (ICLR)}, 2016.

\bibitem{finn}
C. Finn, P. Abbeel, and S. Levine,
\newblock ``Model-agnostic Meta-learning for Fast Adaptation of Deep Networks,''
\newblock {\em In Proceedings of the 34th International Conference on Machine Learning-Volume 70}, vol. 100, pp. 1126-1135,  2017.

\bibitem{12}
R. Chattopadhyay, N. C. Krishnan, and S. Panchanathan,
\newblock ``Topology Preserving Domain Adaptation for Addressing Subject based Variability in SEMG Signal,''
\newblock {\em in Proc. AAAI Spring Symp., Comput. Physiol.}, 2011, pp. 4–9.

\bibitem{transfer-learning}
U. Côté-Allard, C.L. Fall, A. Drouin, A. Campeau-Lecours, C. Gosselin, K. Glette, F.  Laviolette, and B. Gosselin,
\newblock ``Deep Learning for Electromyographic Hand Gesture Signal Classification using Transfer Learning,''
\newblock {\em IEEE Trans. Neural Syst. Rehabil. Eng.}, vol. 27, no. 4, pp. 760-771, 2019.

\bibitem{transfer-learning2}
U. Côté-Allard, C. L. Fall, A. Campeau-Lecours, C. Gosselin, F. Laviolette, and B. Gosselin,
\newblock ``Transfer learning for SEMG Hand Gestures Recognition using Convolutional Neural Networks,''
\newblock {\em Proc. IEEE Int. Conf. Syst., Man, Cybern.},  Oct. 2017, pp. 1663–1668.

\bibitem{4_Dario}
M. Zia ur Rehman, A. Waris, S.O. Gilani, M. Jochumsen, I.K. Niazi, M. Jamil, D. Farina, and  E.N. Kamavuako,
\newblock ``Multiday EMG-based Classification of Hand Motions with Deep Learning Techniques,''
\newblock {\em Sensors}, vol. 18, no. 8, p. 2497,  2018.

\bibitem{inter-session}
Y. Du, W. Jin, W. Wei, Y. Hu, and W. Geng,
\newblock ``Surface Emg-based Inter-session Gesture Recognition Enhanced by Deep Domain Adaptation,''
\newblock {\em Sensors}, vol. 17, no. 3, p. 458,  2017.

\bibitem{24}
B. Hudgins, P. Parker, and R.N. Scott,
\newblock ``A New Strategy for Multifunction Myoelectric Control,''
\newblock {\em IEEE Trans. Biomed. Eng.},  vol. 40, no. 1, p.82-94, 1993.

\bibitem{10}
S. Bai, J.Z. Kolter, and V. Koltun,
\newblock ``An Empirical Evaluation of Generic Convolutional and Recurrent Networks for Sequence Modeling,''
\newblock {\em arXiv preprint arXiv:1803.01271},2018.

\bibitem{11}
A.V.D. Oord, \textit{at al.},
\newblock ``Wavenet: A Generative Model for Raw Audio,''
\newblock {\em ArXiv preprint arXiv:1609.03499}, 2016.

\bibitem{Vaswani}
A. Vaswani, N. Shazeer, J. Uszkoreit, L. Jones, A. Gomez N., L. Kaiser, and
I. Polosukhin,
\newblock ``Attention is All You Need,''
\newblock {\em  arXiv preprint arXiv:1706.03762}, 2017a.\\

\bibitem{Elaheh:Code}
https://ellarahimian.github.io/FS-HGR/\\

\end{thebibliography}

\end{document}